%% file: main.tex
\documentclass[sigconf]{acmart}

\AtBeginDocument{%
  \providecommand\BibTeX{{%
    \normalfont B\kern-0.5em{\scshape i\kern-0.25em b}\kern-0.8em\TeX}}}

\copyrightyear{2021}
\acmYear{2021}
\setcopyright{acmcopyright}\acmConference[AIES '21]{Proceedings of the 2021
AAAI/ACM Conference on AI, Ethics, and Society}{May 19--21, 2021}{Virtual Event,
USA}
\acmBooktitle{Proceedings of the 2021 AAAI/ACM Conference on AI, Ethics, and
Society (AIES '21), May 19--21, 2021, Virtual Event, USA}
\acmPrice{15.00}
\acmDOI{10.1145/3461702.3462587}
\acmISBN{978-1-4503-8473-5/21/05}

\usepackage{graphicx} 
\usepackage{natbib}  
\usepackage{caption} 

\usepackage{siunitx}
\usepackage{natbib}
\usepackage{multirow}
\usepackage{amsmath}
\usepackage{subdepth}
\usepackage{graphicx}
\usepackage{subcaption}
\usepackage{booktabs} 
\usepackage{multirow} 
\usepackage{algorithmic}
\usepackage{algorithm}
\usepackage{tikz}
\usetikzlibrary{shapes,decorations,arrows,calc,arrows.meta,fit,positioning}
\tikzset{
    -Latex,auto,node distance =0.6 cm and 0.6 cm, thick, line width = 1.5,
    state/.style ={circle, draw, thick, minimum width = 0.8 cm, line width=1pt},
    point/.style = {circle, draw, inner sep=0.04cm,fill,node contents={}},
    bidirected/.style={Latex-Latex,dashed},
    el/.style = {inner sep=2pt, align=left, sloped},
    every picture/.style={line width=1pt}
}

\DeclareMathOperator*{\argmin}{arg\,min}

\usepackage{amsfonts}

\global\long\def\ep{\mathbb{E}}
\global\long\def\pr{\text{Pr}}
\global\long\def\cn{\mathcal{N}}

\global\long\def\pr{\text{Pr}}

\global\long\def\ep{\mathbb{E}}
\global\long\def\cn{\mathcal{N}}

\global\long\def\pa{\mathit{pa}}

\global\long\def\ep{\mathbb{E}}
\global\long\def\pr{\text{Pr}}
\global\long\def\cn{\mathcal{N}}

\newtheorem{prop}{Proposition}

\newtheorem{definition}{Definition}[section]

\begin{document}

\title{Causal Multi-Level Fairness}

\author{Vishwali Mhasawade}
\email{vishwalim@nyu.edu}
\affiliation{%
  \institution{Tandon School of Engineering\\ New York University}
  \state{New York}
  \country{USA}
}
\author{Rumi Chunara}
\email{rumi.chunara@nyu.edu}
\affiliation{%
  \institution{Tandon School of Engineering;\\School of Global Public Health\\New York University}
  \state{New York}
  \country{USA}
  }

\fancyhead{} 

\begin{abstract}
Algorithmic systems are known to impact marginalized groups severely, and more so, if all sources of bias are not considered. While work in algorithmic fairness to-date has primarily focused on addressing discrimination due to individually linked attributes, social science research elucidates how some properties we link to individuals can be conceptualized as having causes at macro (e.g. structural) levels, and it may be important to be fair to attributes at multiple levels. For example, instead of simply considering race as a causal, protected attribute of an individual, the cause may be distilled as perceived racial discrimination an individual experiences, which in turn can be affected by neighborhood-level factors. This multi-level conceptualization is relevant to questions of fairness, as it may not only be important to take into account if the individual belonged to another demographic group, but also if the individual received advantaged treatment at the macro-level. In this paper, we formalize the problem of multi-level fairness using tools from causal inference in a manner that allows one to assess and account for effects of sensitive attributes at multiple levels. We show importance of the problem by illustrating residual unfairness if macro-level sensitive attributes are not accounted for, or included without accounting for their multi-level nature. Further, in the context of a real-world task of predicting income based on macro and individual-level attributes, we demonstrate an approach for mitigating unfairness, a result of multi-level sensitive attributes.
\end{abstract}



\maketitle
\input{introduction}

\input{related_work}

\input{background}
\input{multi-level_fairness}

\input{experiments}

\input{discussion}

\section*{Acknowledgments}
We acknowledge funding from the National Science Foundation, award number 1845487. We also thank Harvineet Singh for helpful discussions.

\bibliographystyle{ACM-Reference-Format}
\bibliography{aies_ref}

\end{document}


\maketitle
\appendix
\input{appendix}
\bibliography{aies_ref.bib}

%% file: introduction.tex
\section{Introduction}
There has been much recent interest in designing algorithms that make fair predictions \citep{kroll2016accountable,diakopoulos2017principles,barocas2017problem}. Definitions of algorithmic fairness have been summarized in detail elsewhere \citep{verma2018fairness,mishler2020fairness}; broadly approaches to algorithmic fairness in machine learning can be divided into two kinds: group fairness, which ensures some form of statistical parity for members of different protected groups \citep{dwork2012fairness,mishler2020fairness} and individual notions of fairness which aim to ensure that people who are ‘similar’ with respect to the classification task receive similar outcomes \citep{binns2020apparent,joseph2016rawlsian,biega2018equity}. Historically sensitive variables such as age, gender, race have been thought of as attributes that can lead to unfairness or discrimination. Recently, the causal framework \citep{pearl2000models} has also been used to conceptualize fairness. \citet{kusner2017counterfactual} introduced a causal definition of fairness, \textit{counterfactual fairness}, which states that a decision is fair toward an individual if it coincides with the one that would have been taken in a counterfactual world in which the sensitive attribute of the individual had been different. This approach assumes prior knowledge about how data is generated and how variables affect each other, represented by a causal diagram and considers that all paths from the sensitive attribute to the outcome are unfair. Counterfactual fairness, thus, postulates that the entire effect of the sensitive attribute on the decision, along all causal paths from the sensitive attribute to the outcome is unfair. This approach is restrictive when only certain causal paths are unfair. For example, the effect of gender on income levels can be considered unfair but the effect through education attainment level is fair \citep{chiappa2019path}. Path-specific counterfactual fairness has then been conceptualized, which attempts to correct the effect of the sensitive attribute on the outcomes \emph{only} along unfair causal pathways instead of entirely removing the effect of the sensitive attribute \citep{chiappa2019path}. It should be noted that throughout this work, determination of fair and unfair paths requires domain expertise and possibly discussions between policymakers, lawyers, and computer scientists. 

Leveraging domain knowledge from sociology highlights that notwithstanding these clarifications of path-specific fairness, algorithmic fairness efforts in general have been limited to sensitive attributes at only the \emph{individual-level}, missing important assertions regarding how social aspects can be influenced and created more by societal structures and cultural assumptions than by individual and psychological factors \citep{kincheloe2011rethinking}. Critical Theory is an approach in sociology which motivates the consideration of macro-structural attributes, often simply called macro-properties or ‘structural attributes’. These structural attributes correspond to overall organizations of society that extend beyond the immediate milieu of individual attributes yet have an impact on an individual, such as social groups, organizations, institutions, nation-states, and their respective properties and relations \cite{furze2014sociology,little2016introduction}. Accounting for a multi-level conceptualization in algorithmic fairness is consistent with recent literature which have discussed similar concerns with the conceptualization of race as a fixed, individual-level attribute \citep{vanderweele2014causal,hanna2020towards,zuberi2000deracializing,boyd2020racism}. Analytically framing race as an individual-level attribute ignores systemic racism and other factors which are the mechanisms by which race has consequences \citep{boyd2020racism}. Instead, accounting for the macro-level phenomena associated with race, which a multi-level perspective promotes (e.g. structural racism, childhood experiences that shape the perception of race) will augment the impact of algorithmic fairness work \citep{hanna2020towards}. Several different terms have been used as synonyms for macro-structural attributes, including ``group-level'' variables in multi-level analysis, ``ecological variables'', ``macro-level variables'', ``contextual variables'', ``group variables'', and ``population variables'', and we refer the reader to a full elaboration on macro-property variables in previous work by \citet{diez1998bringing}. In this paper, we refer to all information measured beyond an individual-level as \textit{macro-properties} and \textit{macro-level} attributes. 

Without accounting for macro-properties, assessment of how ``fair'' an algorithm is may inadvertently be unfair by not accounting for important variance in attributes that are at the structural level. Indeed, today's growing availability of data that captures such effects means that leveraging the right data can be used to account for and address unexplained variance when using individual-level attributes as proxies for structural attributes. This can help work towards equity versus in-sample fairness (e.g. an algorithm fair with respect to perceived race will still be unfair to structural differences that perceived race may be a proxy for, such as neighborhood socio-economics, and including the structural factors can help to mitigate unfairness within the perceived race category) \citep{mhasawade2020machine}. Accounting for structural factors can also clearly identify populations that input data is not representing. Indeed, many macro-properties can be considered as sensitive attributes. For example, neighborhood socioeconomic status which is a measure of the inequality in distribution of macro-level resources of education, work, and economic resources\citep{ross2008neighborhood}. It has been shown that patients who reside in low socioeconomic neighborhoods can have significantly higher risk for readmission following hospitalization for sepsis in comparison to patients residing in higher socioeconomic neighborhoods even if all individual-level risk factors are the same, indicating that low SES is an independent factor of relevance to the outcome \citep{galiatsatos2020association}. Thus, accounting for relevant macro-level properties while making decisions, say,  about post-hospitalization care, is critical to ensure that disadvantaged patients, comprehensively accounting for all relevant sources of unfairness, receive appropriate care.

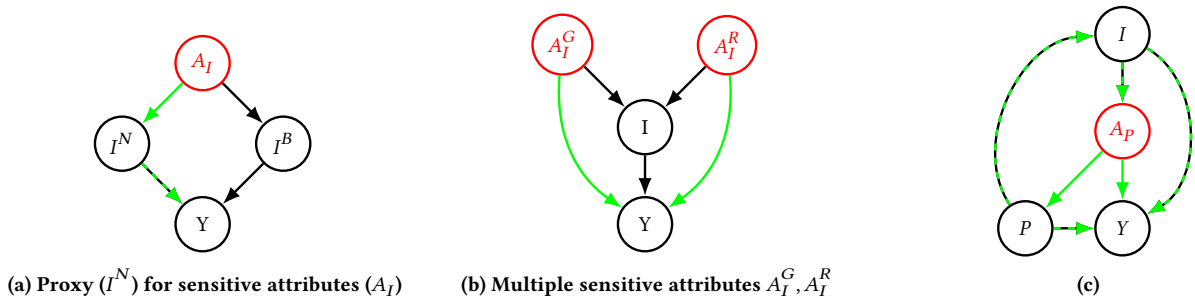
\begin{figure*}
\centering
\begin{subfigure}[b]{.33\textwidth}
    \centering
    \scalebox{0.9}{\input{Figures/proxy}}
    \caption{Proxy ($I^N$)  for sensitive attributes ($A_I$)}
    \label{fig:proxy}
\end{subfigure}%
\begin{subfigure}[b]{.33\textwidth}
    \centering
    \scalebox{0.9}{\input{Figures/multiple_sensitiv}}
    \caption{Multiple sensitive attributes $A_I^G,A_I^R$}
    \label{fig:multi_sensitive}
\end{subfigure}%
\begin{subfigure}[b]{.33\textwidth}
    \centering
        \scalebox{0.9}{\input{Figures/failure_case}}
    \caption{}
    \label{fig:failure}
\end{subfigure}%
    \caption{Causal graphs to describe and distinguish different general sensitive attribute settings. Green arrows denote discriminatory causal paths and dashed green-black arrows from any node $R$ to $S$ denote discrimination only due to the portion of the effect from a green arrow into $R$ and not from $R$ itself (fair paths defined via a priori knowledge), red nodes represent sensitive attributes. a) Individual-level variable $I^N$ (e.g. name) acts as a proxy for the individual sensitive attribute $A_I$ (e.g. perceived race) and affects outcome $Y$ (e.g. health outcomes). $I^B$ (e.g. a biological factor) is not a proxy but also affects the outcome. b) Multiple sensitive attributes at the individual level, $A_I^G$ (e.g. gender) and $A_I^R$ (e.g. perceived race) affect individual variables, $I$, and health outcome, $Y$. c) Example of cycles in the causal graph, where individual level attribute $I$, individual income, causes macro-level sensitive attribute, mean neighborhood income, $A_P$, which in turn affects education resources in the neighborhood, $P$, with $Y$ being the outcome. Since $I \in pa_{A_P}$, this violates the necessary conditions for obtaining a fair predictor with multi-level causal fairness.  }
    \label{fig:interactions}
\end{figure*}

Towards this goal, we propose a novel definition of fairness called causal \emph{multi-level fairness}, which is defined as a decision being fair towards an individual if it coincides with the one in the counterfactual world where, contrary to what is observed, the individual receives advantaged treatment at both the \emph{macro} and \emph{individual levels}, described by macro and individual-level sensitive attributes, respectively. This work builds upon previous work on path-specific counterfactual fairness, limited to individual-level sensitive attributes, to account for both macro and individual-level discrimination. Multi-level sensitive attributes are a specific case of multiple sensitive attributes, in which we specifically consider a sensitive attribute at a macro level that influences one at an individual level. This is an important and broad class of settings; such multi-level interactions are also considered in multi-level modelling in statistics \citep{gelman2006multilevel}. Moreover, this work addresses a different problem than the setting of proxy variables, in which the aim is to eliminate influence of the proxy on outcome prediction \citep{kilbertus2017avoiding}. Indeed, for variables such as race, detailed analyses from epidemiology have articulated how concepts such as racial inequality can be decomposed into the portion that would be eliminated by equalizing adult socioeconomic status across racial groups, and a portion of the inequality that would remain even if adult socioeconomic status across racial groups were equalized (i.e. racism). Thus, if we simply ascribe the race variable to the individual (as is sometimes done as a proxy), we will miss the population-level factors that affect it and any particular outcome \citep{vanderweele2014causal}.

Explicitly accounting for multi-level factors enables us to decompose concepts such as racial inequality and approach questions such as: what would the outcome be if there was a different treatment on a population-level attribute such as neighborhood socioeconomic status? This is an important step in auditing not just the sources of bias that can lead to discriminatory outcomes but also in identifying and assessing the level of impact of different causal pathways that contribute to unfairness. Given that in some subject areas, the most effective interventions are at the macro level (e.g. in health the largest opportunity for decreasing incidence due to several diseases lie with the social determinants of health such as availability of resources versus individual-level attributes \citep{mhasawade2020machine}), it is critical to have a framework to assess these multiple sources of unfairness. Moreover, by including macro-level factors and their influence on individual ones, we engage themes of inequality and power by specifying attributes outside of an individual's control. In sum, our work is one step towards integrating perspectives that articulate the multi-dimensionality of sensitive attributes (for example race, and other social constructs) into algorithmic approaches. We do this by bringing focus to social processes which often affect individual attributes, and developing a framework to account for multiple casual pathways of unfairness amongst them.  Our specific contributions are:
\begin{itemize}
    \item We formalize multi-level causal systems, which include potentially fair and unfair path effects at both macro and individual-level sensitive variables. Such a framework enables the algorithmicist to conceptualize systems that include the systemic factors that shape outcomes.
    \item Using the above framework, we demonstrate that residual fairness can result if macro-level attributes are not accounted for.
    \item We provide necessary conditions for achieving multi-level fairness and demonstrate an approach for mitigating multi-level unfairness while retaining model performance.
\end{itemize}

%% file: Figures/proxy.tex
\begin{tikzpicture}
    \node[state,red] (a_i) at (0,0) {$A_I$};
    \node[state] (i) [below left =of a_i] {$I^N$};
    \node[state] (i_1) [below right =of a_i] {$I^B$};
    \node[state] (y) [below right =of i] {Y};

    \path [green](a_i) edge (i);
    \path (a_i) edge (i_1);
    \path (i) edge (y);
    \path (i_1) edge (y);
    \path [green, dashed] (i) edge (y);
   
\end{tikzpicture}

%% file: Figures/multiple_sensitiv.tex



\begin{tikzpicture}
    \node[state,red] (a_i) at (0,0) {$A_I^{G}$};
    \node[state] (i) [below right=of a_i] {I};
    \node[state,red] (a_i_2) [above right=of i] {$A_I^ {R}$};
    \node[state] (y) [below =of i] {Y};
    
    \path (a_i) edge (i);
    \path (a_i_2) edge (i);
    \path [green] (a_i) edge [bend right=30] (y);
    \path [green] (a_i_2) edge [bend left=30] (y);
    \path (i) edge (y);
\end{tikzpicture}

%% file: Figures/failure_case.tex
\begin{tikzpicture}
    \node[state] (i) at (0,0) {$I$};
    \node[state,red] (n_i) [below =of i] {$A_P$};
    \node[state] (y) [below =of n_i] {$Y$};
    \node[state] (e_n) [left =of y] {$P$};
    \path (i) edge (n_i);
        \path (i) [green, dashed] edge (n_i);

    \path (n_i) [green] edge (y);
    \path (e_n) edge (y);
    \path (e_n) [green, dashed] edge (y);

    \path (n_i) [green] edge (e_n);
    \path  (e_n)  edge[bend left=60] (i);
    \path  (e_n) [green, dashed] edge[bend left=60] (i);

    \path (i)  edge[bend left=60] (y);
    \path (i) [green, dashed] edge[bend left=60] (y);


\end{tikzpicture}

%% file: related_work.tex
\section{Related Work} 

\textbf{Causal Fairness}.
Several statistical fairness criteria have been introduced over the last decade, to ensure models are fair with respect to group fairness metrics or individual fairness metrics. However, further discussions have highlighted that several of the fairness metrics cannot be concurrently satisfied on the same data \citep{kleinberg2016inherent,chouldechova2017fair,friedler2016possibility,berk2018fairness,kasy2020fairness,mccradden2020ethical}. In light of this, causal approaches to fairness have been recently developed to provide a more intuitive reasoning corresponding to domain specifics of the applications \citep{kilbertus2017avoiding,kusner2017counterfactual,kusner2019making,bonchi2017exposing,qureshi2016causal,russell2017worlds,kusner2018causal,zhang2018equality,zhang2018fairness,nabi2018fair,nabi2019optimal,chiappa2019path}. Most of these approaches advocate for fairness by addressing an unfair causal effect of the sensitive attribute on the decision.  \citet{kusner2017counterfactual} have introduced an individual-level causal definition of fairness, known as counterfactual fairness. The intuition is that the decision is fair if it coincides with the one that would have been taken in a counterfactual world in which the individual would be identified by a different sensitive attribute. For example, a hiring decision is counterfactually fair if the individual identified by the gender \emph{male} would have also been offered the job had the individual identified as \emph{female}. In sum, these efforts develop concepts of fairness with respect to individual level sensitive attributes, while here we develop fairness accounting for multi-level sensitive attributes, comprising of both individual and macro-level sensitive attributes. Inspired from research in social sciences \citep{vanderweele2014causal,hanna2020towards,binns2020apparent}, the work here extends the idea to multi-level sensitive attributes. In our considered setting, a decision is fair not only if the individual identified with a different individual-level sensitive attribute but also if the individual received advantaged treatment at the macro-level, attributed as the macro-level sensitive attribute.

\textbf{Identification of Causal Effects}.
While majority of the work in causal inference is on identification and estimation of the total causal effect \citep{tian2002general,avin2005identifiability,pearl2001,pearl2000models,peters2017elements}, studies have also looked at identifying the causal effect along certain causal pathways \citep{shpitser2013counterfactual,pearl2013direct}. The most common approach for identifying the causal effect along different causal pathways is decomposing the total causal effect along direct and indirect pathways \citep{pearl2013direct,pearl2012causal,shpitser2013counterfactual}. We leverage the approach developed by \citet{shpitser2013counterfactual} on how causal effects of multiple variables along a single causal pathway can be identified. While we assume no unmeasured confounding for the analysis in this work, research in causal estimation in the presence of unmeasured confounding, \citep{miao2018identifying,wang2019multiple} can be used to extend our current contribution.

 \textbf{Path-specific causal fairness}.
Approaches in causal fairness such as path-specific counterfactual fairness \citep{nabi2018fair,chiappa2019path,nabi2019optimal}, proxy discrimination, and unresolved discrimination \citep{kilbertus2017avoiding,wu2019pc} have aimed to understand the effect of sensitive attributes (i.e. variables that correspond to gender, race, disability, or other protected attributes of individuals) on outcomes directly as well as indirectly to identify the causal pathways (represented using a causal diagram) that result into discriminatory predictions of outcomes based on the sensitive attribute. Generally such approaches focus on sensitive attributes at the individual-level and require an understanding of the discriminatory causal pathways for mitigating causal discrimination along the specified pathways. Our approach builds on the work of  \citet{chiappa2019path} for mitigating unfairness by removing path-specific effects for fair predictions, doing so while accounting for both individual and macro-level unfairness.

\textbf{Intersectional fairness}. There is recent focus on identifying the impact of multiple sensitive attributes (intersectionality) on model predictions. There are several works that have been developed for this setting \citep{foulds2020intersectional,yang2020fairness,morina2019auditing}, however, these approaches do not take into account the different causal interactions between sensitive attributes themselves which can be important in identifying bias due to intersectionality. In particular, this includes intersectional attributes at the individual and macro level, such as race and socioeconomic status \citep{coley2015race,kuo2020intersectionality}.

%% file: background.tex
\section{Background}

We begin by introducing the tools needed to outline multi-level fairness, namely, (1) causal models, (2) their graphical definition, (3) causal effects, and (4) counterfactual fairness.\\

\noindent \textbf{Causal Models}: Following \citet{pearl2000models} we define a causal model as a triple of sets $\left(\textbf{U},\textbf{V},F\right)$ such that:
\begin{itemize}
    \item $\textbf{U}$ are a set of latent variables, which are not caused by any of the observed variables in $\textbf{V}$.
    \item $F$ is set of functions for each $V_i \in \textbf{V}$, such that \\ $V_i = f_i\left(\pa_i, U_{pa_i}\right)$, $pa_i\subseteq \textbf{V} \setminus V_i$ and $U_{pa_i} \subseteq \textbf{U}$. Such equations relating $V_i$ to $pa_i$ and $U_{pa_i}$ are known as structural equations \citep{de1993structural}. When $f_i$ is strictly linear, this is known as a linear structural equation model, which is represented as $V_i = \sum_{j=1}^{ \vert pa_i \vert} \alpha_j \cdot pa_i[j] + U_{pa_i}$, where $\alpha$ is the mixing coefficient and $pa_{i}[j]$ denotes the $j$th parent of variable $V_i$.
\end{itemize}

\noindent Here, ``$pa_i$'' refers to the \textit{causal parents} of $V_i$, i.e. the variables that affect the specific value $V_i$ obtains. The joint distribution over all the $n$ variables, $\pr\left(V_1,V_2,\cdots,V_n\right)$ is given by the product of the conditional distribution of each variable, $V_i$, given it's causal parents $pa_i$ as follows:
\begin{align}
    \pr \left(\textbf{V}\right) = \prod_i \pr\left(V_i \vert \pa_i\right).
\end{align}

\noindent \textbf{Graphical definition of causal models}: 
While structural equations define the relations between variables we can graphically represent the causal relationships between random variables using Graphical Causal Models (GCMs) \citep{chiappa2019path}. The nodes in a GCM represent the random variables of interest, $V$, and the edges represent the causal and statistical relationships between them. Here, we restrict our analysis to Directed Acyclic Graphs (DAGs) where there are no cycles, i.e., a node cannot have an edge both originating from and terminating at itself. Furthermore,  a node $Y$ is known as a child of another node $X$ if there is an edge between the two that originates at $X$ and terminates on $Y$. Here, $X$ is called a direct cause of $Y$. If $Y$ is a descendant of $X$, with some other variable $Z$ along the path from $X$ to $Y$, $X \rightarrow \cdots \rightarrow Z \rightarrow \cdots Y$, then $Z$ is known as a \emph{mediator} variable and $X$ remains as a potential cause of $Y$. For example, in Figure \ref{fig:multi_sensitive},  $A_I^G, A_I^R, I, Y$ are the random variables of interest. $A_I^G, A_I^R$ are direct causes of $I$ and $Y$ while $I$ is a direct cause of $Y$. Here, $I$ is known as a \emph{mediating variable} for both $A_I^G$ and $A_I^R$.\\

 

\noindent \textbf{Causal effects:}
The causal effect of $X$ on $Y$ is the information propagated by $X$ towards $Y$ via the causal directed paths, $X \rightarrow \cdots \rightarrow Y$. This is equal to the conditional distribution of $Y$ given $X$ if there are no bidirected paths between $X$ and $Y$. A bidirected path between $X$ and $Y$, $X \leftarrow\cdots\rightarrow Y$ represents confounding; that is some causal variable known as a confounder, which may be unobserved, affecting both $X$ and $Y$. If confounders are present then the causal effect can be estimated by intervening upon $X$. This means that we externally set the value of $X$ to the desired value $x$ and remove any edges that terminate at $X$, since manually setting $X$ to $x$ would inhibit any causation by $pa_x$. After intervening the causal effect of $X$ on $Y$ is given by $\pr^*\left(Y \vert X = x\right) = \sum_Z \pr\left(Y\vert X=x,Z\right) \pr\left(Z\right)$, where $Z=\mathbf{V}\setminus \{X,Y\}$. The intervention $X=x$ results into a potential outcome, $Y_{X=x}$ (also represented as $Y_x$), where the distribution of the potential outcome variable is defined as $\pr\left(Y_{x}\right) = \pr^*\left(Y \vert X=x\right)$.\\

\noindent \textbf{Counterfactual fairness:}
Counterfactual fairness is a causal notion of fairness that restricts the decisions of the algorithm to be invariant to the value assigned to the sensitive variable. Assuming $A$ to be the sensitive attribute obtaining only two values, $a$ and $a'$, $X$ as the remaining variables, and $Y$ as the outcome of interest, counterfactual fairness is defined as follows:
\begin{definition}
Predictor $\hat{Y}$ is \textbf{counterfactually fair} if under any context $X$ and $A=a$,\end{definition}
\begin{align}\small
 \pr(\hat{Y}_{A\rightarrow a}(U)=y \vert X,A=a) = \pr(\hat{Y}_{A\rightarrow a'}(U)=y \vert X,A=a) 
\end{align}
With this background, we next discuss the problem setup.


\section{Problem Setup}
Let us denote all the variables associated with the system being modelled as $\textbf{V}:= \{A_I, A_P, \textbf{I}, \textbf{P}, Y\}$, where $A_I$ and $A_P$ are the sensitive attributes at the individual-level and the macro-level respectively\footnote{Variables will be denoted by uppercase letters, $V$, values by lowercase letters, $v$, and
sets by bold letters, $\mathbf{V}$.}, $\textbf{I}$ is a non empty set of individual level variables, $\textbf{P}$ is a non empty set of macro-level variables, and $Y$ is an outcome of interest. We assume access to a causal graph, $\mathcal{G}$, consisting of $\textbf{V}$ which accurately represents the data-generating process. Our goal is to obtain a classifier, $\hat{Y}_{\text{fair}}$, which is fair with respect to both $A_P$ and $A_I$. While previous approaches in algorithmic fairness have only considered $A_I$ as a sensitive attribute, we present a novel approach to also counter any discrimination resulting from $A_P$. While macro-level attributes, $\mathbf{P}$ and $A_P$, may seem like regular nodes in the causal graph, the difference is that based on them being macro-properties, their effects on the outcome may be mediated via the individual level attributes, $\mathbf{I}$ and $A_I$, making it nontrivial to mitigate unfairness due to these macro-properties. We  highlight that this paper presents, to our knowledge, the first approach for incorporating unfairness due to macro-properties into algorithmic fairness. In order to further help guide the reader regarding differences in macro and individual-level variables, we clarify the types of macro-level attributes which are considered in our setting. Per the extant literature \citep{diez1998bringing}, macro-level attributes can be categorized into two types:\\
\begin{itemize}
    \item \textbf{Category 1}: non-aggregate group properties like nationality, existence of certain types of regulations, population density, degree of income inequality in a community, political regime, legal status of women which do not include individual properties in their computation
    \item \textbf{Category 2}: aggregate properties such as neighborhood income and median household income, which include individual properties (here individual income) in their computation
\end{itemize}
   Thus, individual variables influence aggregate macro-properties (Category 2) but not non-aggregate macro-properties (Category 1). This categorization of macro-properties helps us to identify the necessary conditions for obtaining a fair classifier with respect to both individual and macro-level sensitive attributes. In parallel, a main assumption in causal fairness and causal inference settings is that we have access to a causal graph and that the graph does not include any bidirected edges or cycles,   \citep{chiappa2019path,nabi2018fair,kusner2017counterfactual} which are assumptions we maintain for the causal multi-level fairness setting. This assumption also clarifies that we should limit our consideration of macro-level attributes to Category 1 and not consider aggregate information like mean neighborhood income as macro-level sensitive variables since individual-level information has a causal influence on mean neighborhood income, i.e. $I \rightarrow A_P$. Certainly, an increase in the individual income increases mean neighborhood income, which can result in feedback loops in the causal graph which are disallowed. Figure \ref{fig:failure} represents such a case involving cycles which the current framework does not support. Further, we assume that the structural equation model representing the functional form underlying the causal graph is linear in nature, i.e. a node is a linear combination of its causal parents. Future work should extend the analysis to non-linear relationships and non-parametric forms. We also assume that we are aware about the fair and unfair causal paths for the multi-level sensitive attributes, again consistent with previous work \citep{chiappa2019path}. Next, we present an example to realize the importance of macro-level sensitive attributes.

\subsection{An Illustrative Example}
In prior sections we have motivated the need to consider macro-level attributes. We now describe multi-level sensitive attributes and provide an illustrative example of why considering macro-level sensitive attributes in algorithmic fairness efforts can be important. Consider neighborhood socioeconomic status (SES) as $A_P$. Neighborhood SES can be a cause of population-level factors such as resources available. Neighborhood SES can also affect individual-level attributes, $A_I$ such as perception of race\footnote{We note that race is a social construct and henceforth use ``perceived racial discrimination'' here consistent with best practice \cite{adams2007teaching}.} (which is protected), by shaping the cultural and social experiences of an individual \citep{dailey2010neighborhood} as well as non-protected attributes, $I$ (e.g. body-mass index, given that neighborhood SES can shape the types of resources available and norms \citep{mathur2013individual}). Furthermore, research has described how macro attributes such as neighborhood SES can affect health behaviors like smoking directly as well as through individual factors such as perceived racial discrimination. Thus, given that this macro-attribute shapes the resources and circumstances of an individual ($A_I$ can be influenced by $A_P$), it could be desirable for algorithms which are being used to decide fair allocation of health services or resources to predicted smokers, to account for this macro-property. Not being fair to this property lacks consideration for variance within the individual attributes that it influences. In other words, accounting for only individual-level attributes such as perceived racial discrimination can still lead to unfairness without considering that people within the same individual-level category could have different resources or opportunities available to them.

A general framework for macro/individual attribute relationships and possible unfair paths in presented in Figure \ref{fig:multi_level}. In illustrating the relationship between $A_P$ and $A_I$ (green arrow) we make distinction between the multi-level fairness setting and literature on intersectionality in fairness, which has considered multiple sensitive attributes that are independent of each other (Figure \ref{fig:multi_sensitive}) \citep{foulds2020intersectional,morina2019auditing,yang2020fairness,yang2020causal}. While this assumption of independence may hold for individual-level sensitive attributes like age and gender, it fails when we consider sensitive attributes at both the individual and macro level. It is important to understand the relationship between individual and macro-level attributes in order to delineate the path-specific effects that may lead to biased predictions of $Y$.

Next, we describe how these interactions can lead to unfairness, by introducing multi-level path-specific fairness accounting for both individual and macro-level sensitive attributes.


%% file: multi-level_fairness.tex
\section{Multi-level path-specific fairness}\label{path_specific}
We begin the discussion about multi-level path-specific fairness by first introducing path-specific effects, discussing the necessary conditions to identify the effects with multiple sensitive attributes, and finally describe an approach to obtain fair predictions with multi-level path specific effects.

\begin{figure}[!htbp]
    \centering
    \scalebox{0.9}{\input{Figures/all_paths}}
    \caption{Multi-level sensitive attributes $A_I,A_P$. Macro-level variables, $A_P$ (e.g. neighborhood SES), $P$ (e.g. other zipcode level factors), and individual-level ones, $A_I$ (e.g. perceived racial discrimination), $I$, affect the outcome $Y$ (e.g. a health behavior).}
    \label{fig:multi_level}
\end{figure}
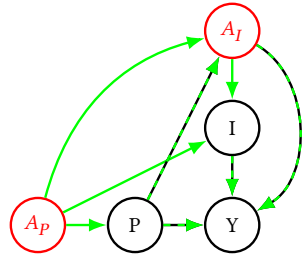

\subsection{Path-specific effects} 
While counterfactual fairness assumes that the entire effect of the sensitive variables on the outcomes is problematic and constructs the predictor with only the non-descendants of the sensitive attribute \citep{kusner2017counterfactual}, it is a fairly restrictive scenario especially when the sensitive attributes can affect the outcome along both \textit{fair} and \textit{unfair} pathways as shown in Figure \ref{fig:multi_level}. Thus, disregarding the effect only along unfair paths can help in preserving the fair individual information regarding sensitive attributes \citep{chiappa2019path}. Indeed, domain experts and policymakers can guide these decisions about identifying fair and unfair pathways. We are interested only in correcting the effect of the sensitive attribute on the outcome along the unfair paths. For this we turn to path-specific effects, which isolate the causal effect of a treatment on an outcome \emph{only} along a certain causal pathway \citep{pearl2001}. 
In order to obtain the causal effect only along a certain pathway, the idea that path-specific effects can be formulated as
nested counterfactuals is leveraged \citep{shpitser2013counterfactual}. This is done by considering that along the causal path of interest, say, $ A_I \rightarrow I \rightarrow Y$, the variable, $A_I$, propagates the observed value, for example, $a_i$ in Figure \ref{fig:multi_level}, and along other pathways, $A_I \rightarrow Y$, the counterfactual value, $a_i'$ is propagated, assuming that $A_I = \{a_i,a_i'\}$. In this case, the effect of any mediator between $A_I$ and $Y$ is evaluated according to the corresponding value of $A_I$ propagated along the specific causal path, the mediator $I$ obtains the value $I(a_i)$.
Here, the primary object of interest is the potential outcome variable, $Y(a_i)$, which represents the outcome if, possibly contrary to fact, $A_I$ were externally set to value $a_i$. Given the values of $a_i,a_i'$, comparison of $Y(a_i')$ and $Y(a_i)$ in expectation: $\ep[Y(a_i)] -\ep[Y(a_i')]$ would allow us to quantify the path-specific causal effect, PSE, of $A_I$ on $Y$.

Using this concept of path-specific effect, we next state path-specific counterfactual fairness, defined by \citet{chiappa2019path}.

\begin{definition}
Path-specific counterfactual fairness is defined as a decision being fair towards an individual if it coincides with the one that would have been made in a counterfactual world in which the sensitive attribute along the unfair pathways (denoted by $\pi$) was set to the counterfactual value.\end{definition} 
\begin{align} \small \pr(\hat{Y}_{A\rightarrow a _ \pi}(U)=y\vert X,A=a) = \pr(\hat{Y}_{A\rightarrow a' _ \pi}(U)=y \vert X,A=a) 
\end{align}
\citet{chiappa2019path} further propose that in order to obtain a fair decision, the path-specific effect of the sensitive attribute along the discriminatory causal paths is removed from the model predictions, i.e. $\hat{Y}_{\text{fair}} = \hat{Y} - \text{PSE}$.  However, a key limitation of the approach presented by \citet{chiappa2019path} is that only the path-specific counterfactual fairness with respect to the sensitive attributes at the individual level is considered. The presence of sensitive attributes at the macro-level that affect properties at the individual level makes it non-trivial to estimate the path-specific effect consisting of both macro and individual-level sensitive attributes. Following, we describe the identifiability conditions for obtaining path-specific counterfactual fairness with respect to multi-level sensitive attributes by estimating multi-level path-specific effects. \\


\noindent \textbf{Identification of multi-level path-specific effects.}
\begin{prop}\label{prop:identification}
In the absence of any unmeasured confounding between $A_P$ and $A_I$ and $\{A_I,\textbf{I}\} \not\in pa_{A_P} $, the multi-level path-specific effects of both $A_P$ and $A_I$ on $Y$ are identifiable.
\end{prop}

Proposition \ref{prop:identification} follows from the possible structural interactions between $A_P$ and $A_I$, where $A_P$ is a non-aggregate macro-property not involving any individual-level computation. Next, we discuss the interaction between $A_P$ and $A_I$, where the macro-level sensitive attribute is a causal parent of the individual-level sensitive attribute, $A_P \in \{pa_{A_I}\}$.

\begin{figure*}[!htbp]
    \centering
    \begin{subfigure}[b]{0.33\linewidth}
    \centering
    \begin{align*} 
\begin{split}
    &A_P \sim \text{Bernoulli}\left(p = 0.5\right),\\
    &P = \theta^p_{\phantom{p}} + \theta^p_{a_p}A_P + \epsilon_p,\\
    &\gamma = \theta^{a_i}+\theta^{a_i}_{a_p} A_P + \theta^{a_i}_{p} P + \epsilon_{a_i},\\
    &A_I = \text{Bernoulli} (\sigma (\gamma)),\\
    &I = \theta^i_{\phantom{p}} + \theta^i_{a_p}A_P + \theta^i_{a_i}A_I + \epsilon_{i},\\
    &Y = \theta^y_{\phantom{p}} + \theta^y_p P + \theta^y_i I +\theta^y_{a_i}A_I +\epsilon_y\\
    & \theta^p, \theta^{a_i},\theta^{i},\theta^{y}= \left(0.2, 0.2, 0.2, 0.2 \right) \\
    & \theta^p_{a_p},\theta^{a_i}_{a_p},\theta^{a_i}_{p},\theta^i_{a_p},\theta^i_{a_i}  \sim U\left(0.2,0.7\right) \\
    & \theta^y_p,\theta^y_{i},\theta^y_{a_i} \sim U\left( 0.3,0.95\right)\\
    & \epsilon_p,\epsilon_{a_i},\epsilon_{i},\epsilon_y \sim \cn\left(0,1\right)\\
    &\sigma = 1/\left(1 + \text{exp}\left(-x\right)\right)
\end{split}
\end{align*}
\caption{}
\label{eq:dgp}
    \end{subfigure}
    \hfill
    \begin{subfigure}[b]{0.33\linewidth}
    \centering
     \scalebox{1}{\input{Figures/pop_as_indi}}   
    
     \vspace{0.7cm}
    \begin{tabular}{|c|c|}
        \hline
         $A_P$& $\theta^y_p  \theta^p_{a_p}$  \\
         \hline
         $A_P,A_I$& $\theta^y_p  \theta^p_{a_p} + \theta^y_{i}  \theta^i_{a_i}$\\
         \hline
    \end{tabular}
    
 
    \caption{}
    \label{fig:ap_as_ai}
    \end{subfigure}%
    \hfill
    \begin{subfigure}[b]{0.33\linewidth}
    \centering
    \includegraphics[scale=0.45,trim=1.5cm 0cm 0cm 0cm]{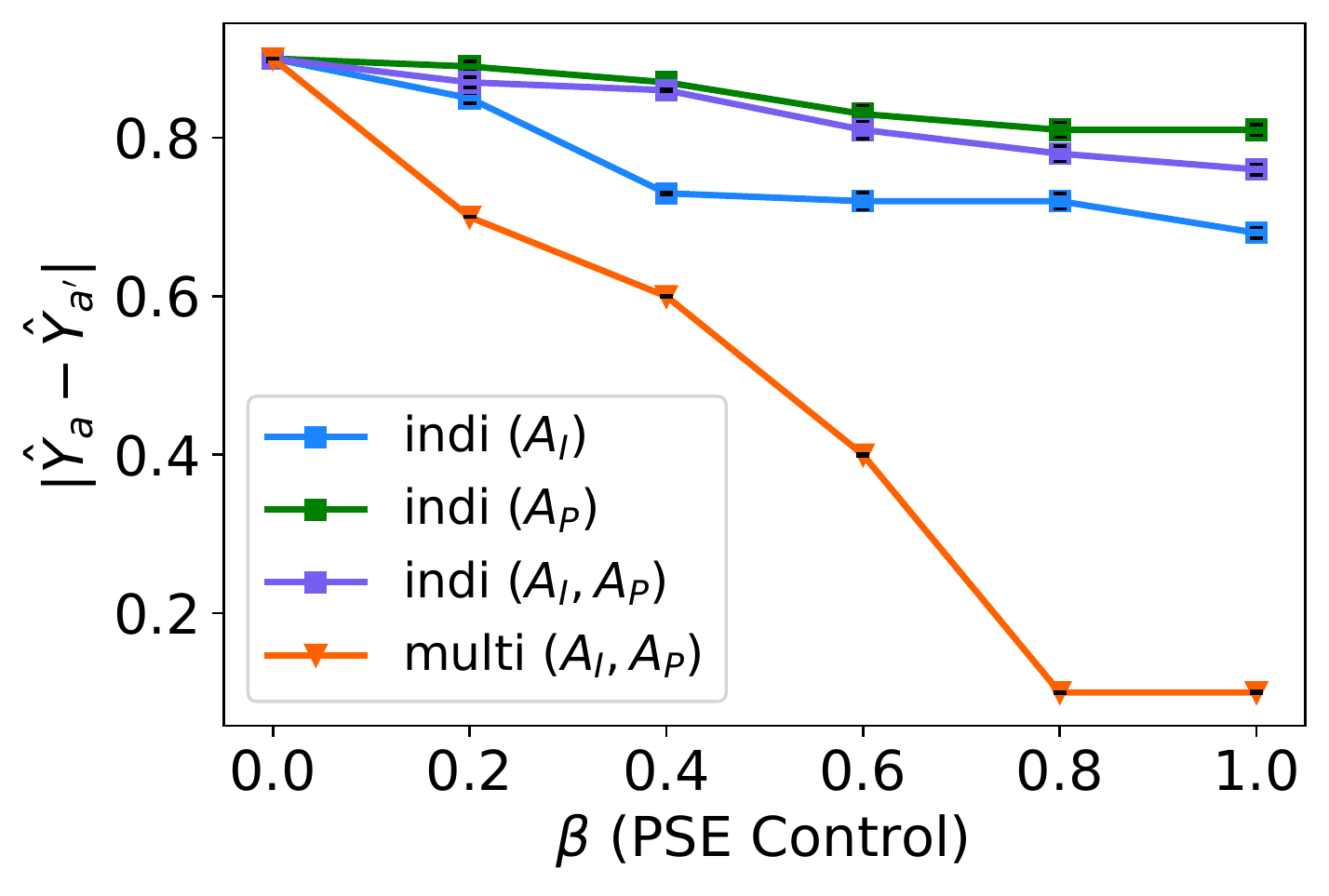}
    \caption{}
    \label{fig:unfairness}
    \end{subfigure}
    \caption{Experimental details regarding multi-level example (Figure \ref{fig:multi_level}), a) Data generating process, b) Resulting causal graph if $A_P$ is considered as an individual-level sensitive attribute instead of macro-level sensitive attribute as presented in Figure \ref{fig:multi_level}, $A_P \not \in pa_{A_I}$, resulting path-specific effects with $A_P$ as individual-level sensitive attribute, and c) Path-specific unfairness controlling for discriminatory effects of just $A_I$ (blue), $A_P$ considered as an individual-level sensitive attribute instead of macro-level sensitive attribute (green), both $A_I,A_P$ considered as individual-level sensitive attributes (purple), and  $A_I,A_P$ considered as multi-level sensitive attributes (orange), over 10 runs.}
\end{figure*}

Figure \ref{fig:multi_level} presents such an exemplar case of multi-level interactions where macro-level sensitive attributes can cause individual level variables, with the data generating process as described in Figure \ref{eq:dgp}. $A_P$ and $A_I$ are binary variables where we assume $a_p'$ and $a_i'$ to represent the baseline values denoting advantaged treatments. The rest of the variables $P,I$ and $Y$ are continuous and follow a linear relationship between the parents and the specific variables. $\epsilon_p, \epsilon_{a_i},\epsilon_i, \epsilon_y$ are unobserved zero-mean Gaussian terms. For this specific model we obtain the multi-level path-specific effects consisting of both $A_P$ and $A_I$. The population level sensitive attribute, $A_P$ affects $Y$ via multiple paths along 1) $A_P \rightarrow P \rightarrow Y$, 2) $A_P \rightarrow I \rightarrow Y$, 3) $A_P \rightarrow A_I \rightarrow Y$, and 4) $A_P \rightarrow A_I \rightarrow I \rightarrow Y$; the individual level sensitive attribute $A_I$ affects $Y$ along two causal paths, 1) $A_I\rightarrow I \rightarrow Y$, and 2) $A_I \rightarrow Y$. With prior assumption that $A_I \rightarrow Y$ and $P\rightarrow Y$ are not discriminatory causal paths, we evaluate the path-specific effect along $\cdots \rightarrow I \rightarrow Y$, where $\cdots \rightarrow I$ repesents all paths into $I$ . Overall the path-specific effect can be calculated as follows:
\begin{align}
    \text{PSE}_{A_P\leftarrow a_p, A_I \leftarrow a_i}^{\cdots \rightarrow I \rightarrow Y} = \ep\left[Y\left(a_i',P\left(a_p'\right),I\left(a_i,a_p\right)\right)\right].
\end{align}

The path-specific effect along $\cdots \rightarrow I \rightarrow Y$ is computed by comparison of the counterfactual variable  $Y\left(a_i',P\left(a_p'\right),I\left(a_i,a_p\right)\right)$, where $a_p,a_i$ are set to the baseline value of 1 in the counterfactual and 0 in the original. The counterfactual distribution can be estimated 
as follows:
\begin{equation}
\begin{aligned}
   &Y\left(a_i',P\left(a_p'\right),I\left(a_i,a_p\right)\right) \\ &= \int_{P,I} \pr\left(Y\vert a_i',I,P\right) \pr\left(I\vert a_i,a_p\right)\pr\left(P\vert a_p'\right).
\end{aligned}
\end{equation}
We obtain the mean of this distribution as follows:
\begin{equation}
\small
\begin{aligned}
\label{eq:mean}
&\ep[Y \vert I,P,a_i'] \\ &=
    \theta^y_{\phantom{i}} + \theta^y_i \theta^i + \theta^y_p \theta^p + \theta^y_{a_i} a_i' + \theta^y_i \theta^i_{a_i}a_i + \theta^y_i\theta^i_{a_p}a_p + \theta^y_p \theta^p_{a_p} a_p'.
\end{aligned}
\end{equation}
$\text{PSE}_{A_P\leftarrow a_p, A_I \leftarrow a_i}^{\cdots \rightarrow I \rightarrow Y}$ is then obtained by comparing the observed and counterfactual value of the mean of the counterfactual distribution, 

\begin{align}
\label{eq:PSE}
\small
\begin{split}
   \ep[Y \vert a_p, a_i]-\ep[Y \vert a_p', a_i'] &= \theta^y_{a_i} \left(a_i' - a_i'\right) + \theta^y_i \theta^i_{a_i}\left(a_i - a_i'\right)\\
   & + \theta^y_i\theta^i_{a_p} \left(a_p-a_p'\right) + \theta^y_p\theta^p_{a_p} \left(a_p' -a_p' \right) \\
   & = \theta^y_i \theta^i_{a_i}\left(a_i - a_i'\right) + \theta^y_i \theta^i_{a_p}\left(a_p - a_p'\right).
\end{split}
\end{align}
Substituting $a_i=1,a_p=1,a_i'=0,a_p'=0$ in Equation \ref{eq:PSE} we obtain the path-specific effect as a function of the parameters $\theta$\footnote{Plug-in estimators can be used to compute the parameters needed for estimating PSE.}:
\begin{align}
    \text{PSE}^{\cdots \rightarrow I \rightarrow Y}_{A_P\leftarrow a_p, A_I \leftarrow a_i} = \theta^y_i \theta^i_{a_i} + \theta^y_i \theta^i_{a_p}.
\end{align}
By the same approach, path-specific effects for individual and macro-level sensitive attributes as well as the multi-level path specific effects for all the causal paths in Figure \ref{fig:multi_level}, are also evaluated. Estimating the multi-level path-specific effects helps in constructing a fair predictor which is described in the following section.\\

\begin{algorithm}
  \caption{Causal Multi-level Fairness}
  \label{alg:multi-level_causal_fairness}
\begin{algorithmic}
  \STATE {\bfseries Input:} Causal Graph $\mathcal{G}$ consisting of nodes $V$, data $\mathcal{D}$ over $V$, discriminatory causal pathways $\pi$, $\beta$.
    \STATE {\bfseries Output:} Fair predictor $\hat{Y}_{\text{fair}}$, model parameters $\theta$  
  \FOR{$V_i \in \textbf{V}$}
  \STATE Estimate $\hat{\theta}^{V_i} \leftarrow \argmin_\theta \sum_{k \in \mathcal{D}}l\left(V_i^{(k)}, f_i\left(pa_i^{(k)}\right)\right)$
  \ENDFOR
  \STATE Calculate path-specific effects, PSE along $\pi$ using $\theta^{\textbf{V}}$
  \STATE return $\hat{Y}_{\text{fair}} = \ep\left[f_Y\left(\theta^{Y}\right)\right] - \beta* \text{PSE}_{\pi}$

\end{algorithmic}
\end{algorithm}

\subsection {Fair predictions with multi-level path-specific effects.} 
Fair outcomes can be estimated by removing the unfair path-specific effect from the prediction (essentially making a correction on all the descendants of the sensitive attributes), $\hat{Y}$ by simply subtracting it, i.e.  $\hat{Y}_{\text{fair}} = \hat{Y} - \text{PSE}$. As done in \citet{chiappa2019path}, removing such unfair information at test time ensures that the resulting decision coincides with the one that would have been taken in a counterfactual world in which the sensitive attribute along the unfair pathways were set to the baseline. 
We leverage this same idea for multi-level causal fairness by generating
fair predictions,  $\hat{y}_{\text{fair}}$, by controlling for the path-specific effects of multi-level sensitive attributes. For example, consider the PSE resulting from intervening upon both $A_P$ and $A_I$ via the path $\dots \rightarrow I \rightarrow Y$, $\text{PSE}_{A_P\leftarrow a_p,A_I \leftarrow a_i}^{\cdots \rightarrow I \rightarrow Y} = \theta^y_i \left(\theta^i_{a_i} + \theta^i_{a_p}\right)$, we can control for the discrimination at test time as follows:
\begin{align}
    \hat{y}^n_{\text{fair}} = \theta^y_{\phantom{a}} + \theta^y_p p^n + \theta^y_i i^n +\theta^y_a a_i^n - \beta\left(\theta^y_i \left(\theta^i_{a_i} + \theta^i_{a_p} \right) \right)
\end{align}
where $\beta$ accounts for the control over the path-specific effect. For our analysis $\beta$ ranges from 0 to 1, where $0$ denotes that we do not account for any path-specific effect while 1 denotes that we remove the entire path-specific effect. Intermediate values allow for only partial removal of the path-specific effect. We study the effect of the control over different path-specific effects on the performance metric of the classifier. The steps are outlined in Algorithm 1.
 We highlight that an accurate causal graph and a linear functional form are key assumptions of this approach.



%% file: Figures/all_paths.tex

\begin{tikzpicture}
    \node[state,red] (a_i) at (0,0) {$A_I$};
    \node[state] (i) [below =of a_i] {I};
    \node[state] (y) [below =of i] {Y};
    \node[state] (p) [left =of y] {P};
    \node[state,red] (a_p) [left =of p] {$A_P$};
    \path [green] (a_p) edge (p);
    \path [green] (a_p) edge (i);
    \path  [green] (a_p) edge[bend left=30] (a_i);
    \path (p) edge (a_i);
    \path [green,dashed] (p) edge (a_i);
    \path (p) edge (y);
    \path [green, dashed] (p) edge (y);
    \path [green] (a_i) edge (i);

    \path  (a_i) edge[bend left=60] (y);
    \path [green, dashed] (a_i) edge[bend left=60] (y);
    \path (i) edge (y);  
    \path [green, dashed] (i) edge (y);

\end{tikzpicture}

%% file: Figures/pop_as_indi.tex
\begin{tikzpicture}
    \node[state] (y) at (0,0) {$Y$};
    \node[state] (p) [above left=of y] {$P$};
    \node[state,red] (a_p) [above =of p] {$A_P$};
    \node[state] (i) [above right =of y] {$I$};

    \node[state,red] (a_i) [above =of i] {$A_I$};
    \path [green] (a_p) edge (p);
    \path (p) edge (y);
    \path [green, dashed] (p) edge (y);
    \path [green] (a_i) edge (i);

    \path  (a_i) edge[bend right=30] (y);
    \path [green, dashed] (a_i) edge[bend right=30] (y);
    \path (i) edge (y);  
    \path [green, dashed] (i) edge (y);

\end{tikzpicture}

%% file: experiments.tex
\section{Experiments}

Our experiments in this section serve to empirically assess and demonstrate residual unfairness when correcting for path-specific effects of multiple sensitive attributes. 
We consider a synthetic setting demonstrated in Figure \ref{fig:multi_level} and real-world setting for income prediction presented in Figure \ref{fig:uci}; in both cases the structural causal model is known a priori as well as the unfair pathways. In each case, we first learn the parameters of the model ($\Theta$) from the observed data based on the known underlying causal graph, $\mathcal{G}$ as illustrated in Algorithm \ref{alg:multi-level_causal_fairness}. We then draw observed and counterfactual samples from the distribution $\theta$ by sampling from the path-specific counterfactual distribution, referred to as \emph{counterfactual distribution}\footnote{Path-specific counterfactual distribution is obtained by altering the original value to the counterfactual value only along the unfair paths (described by green in the causal graph), and retaining the original value along the fair pathways.}.
We estimate fair predictors for both the observed and counterfactual data using Algorithm \ref{alg:multi-level_causal_fairness}. In case of no resultant unfairness, distributions of the outcome estimate for both observed and counterfactual data should coincide. Density plots that do not coincide signifies residual unfairness. 

\subsection{Synthetic setting}

Here we generate data according to the structural equation model presented in Figure \ref{eq:dgp} with linear relationships with the exception that we consider $Y$ to be binary, using a sigmoid function \footnote{The analysis of the path-specific effects remains similar to the one discussed in Section \ref{path_specific}, where the effects are calculated on the odds ratio scale as \citet{nabi2018fair}.}. This represents the data-generating process in accordance with the multi-level setting presented in Figure \ref{fig:multi_level}. The simulation parameters, $\theta$, are non-negative to ensure a positive path-specific effect, i.e. an unfair effect. Furthermore, the parameters are standardized between 0 and 1 to allow for computational efficiency. The range of the parameters in the uniform distribution is simulated to ensure a realistic setting where the counterfactual distribution is diverse, and the differences between the original and counterfactual distribution is significant to evaluate multi-level fairness. If there are little differences in the original and the counterfactual distribution, path-specific effects become negligible resulting in no significant unfairness issues.
In total, 10 simulations with $N=2000$ samples in each are performed. We train a linear regression model for $I,P$, and a logistic regression model for $A_I,Y$ to learn the parameters ($\theta$). We then calculate three path-specific effects as follows: 1) $\text{PSE}_{A_P,A_I}$ accounting for the effect of both $A_P$ and $A_I$, 2) $\text{PSE}_{A_I}$ accounting for the effect of only $A_I$, and 3) $\text{PSE}_{A_P}$ accounting for the effect of only $A_P$. Details about this calculation are presented in Equation \ref{eq:PSE}. 

We observe that controlling for the multi-level path-specific effects of both $A_P$ and $A_I$ leads to little drop in the model performance (accuracy). We also assess the counterfactual fairness of the resulting model after removing the unfair multi-level path-specific effects. This is done by training a classifier for obtaining fair predictors, $\hat{y}_{a_i,a_p}$, on the actual and counterfactual data respectively, where the counterfactual data is obtained by altering the values of the sensitive variables, $A_P$ and $A_I$. This is done by first learning the parameters, $\theta$ from the original data and then altering the values of the sensitive attributes, $A_P$ and $A_I$ and obtaining the values of $P,I$ and $Y$ from these counterfactual values of $A_P, A_I$ and $\theta$.
We assess the counterfactual fairness of the resulting models while accounting for unfairness due to both $A_P, A_I$ and either $A_I$ or $A_P$, in Figure \ref{fig:unfairness}. The Y axis shows the average difference between observed and counterfactual predictions, $|\hat{Y}_a - \hat{Y}_{a'}|$ where $a$ is the observed value of the sensitive attribute and $a'$ is the corresponding counterfactual value. This difference is analyzed for varying control of the path-specific effect ($\text{PSE}$), $\beta$ along the X-axis. In order to assess the advantage of considering the multi-level nature of the data, we compare with the model presented in Figure \ref{fig:ap_as_ai}, where we treat the macro-level sensitive attribute, $A_P$ as another individual-level sensitive attribute, $A_I$, i.e. $A_P \not \in pa_{A_I}$. The \emph{blue} line represents the difference while accounting only for the individual-level path specific effect $A_I$ while the \emph{orange} line represents the multi-level path-specific effect of both $A_P$ and $A_I$. Similarly, the \emph{green} line represents the setting presented in Figure \ref{fig:ap_as_ai}, where $A_P$, is treated as an individual-level sensitive attribute (multi-level nature of its action removed), and \emph{purple} line represents the case where both $A_I$ and $A_P$ are treated as individual-level sensitive attributes. For counterfactually fair models we expect the difference to be close to zero since controlling for the path-specific effects cancels out the counterfactual change. We observe that this happens only while accounting for the multi-level path-specific effects. Thus, accounting for individual-level path-specific effects solely (blue, green, and purple lines) does not result in counterfactually fair predictions as can be seen from the high difference (0.7- 0.9). Moreover, correcting for the unfair multi-level path-specific effects of both $A_P$ and $A_I$ results in counterfactually fair predictions with a slight drop in accuracy from 94.5\% to 91.5\%. 





\subsection{UCI Adult Dataset}
\begin{figure*}[!htbp]
    \centering
    \begin{subfigure}[b]{0.3\textwidth}
    \centering
        \scalebox{0.65}{\input{Figures/adult_uci}}
    \caption{}
    \label{fig:uci}
    \end{subfigure}%
    \begin{subfigure}[b]{0.3\textwidth}
      \centering
    \includegraphics[scale=0.425,trim=1cm 0cm 0cm 0cm]{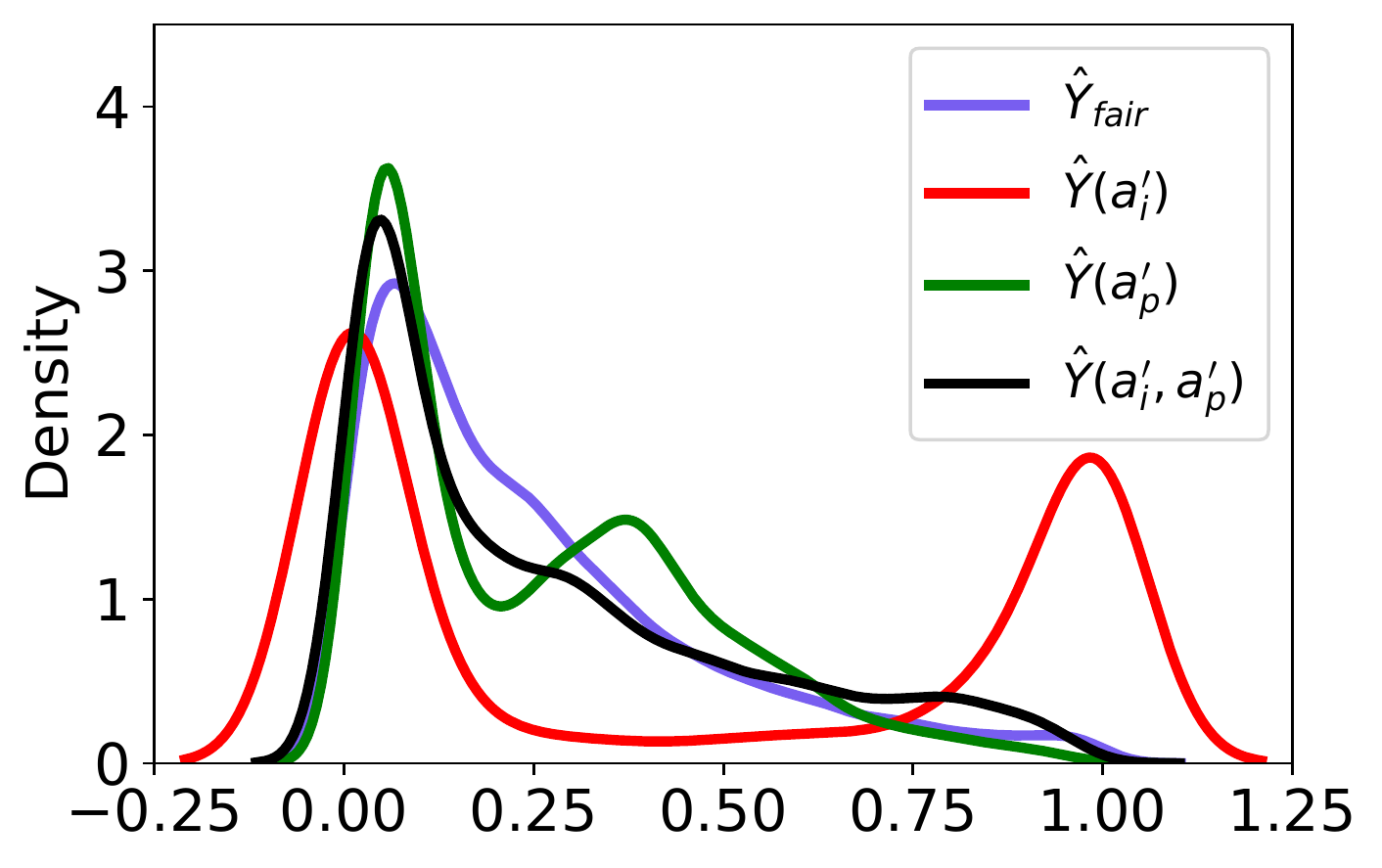}
    \caption{}
    \label{fig:uci_densityplots}
    \end{subfigure}
    \hfill
    \begin{subfigure}[b]{0.3\textwidth}
    \centering
    \includegraphics[scale=0.42,trim=0cm 0.5cm 0cm 0cm]{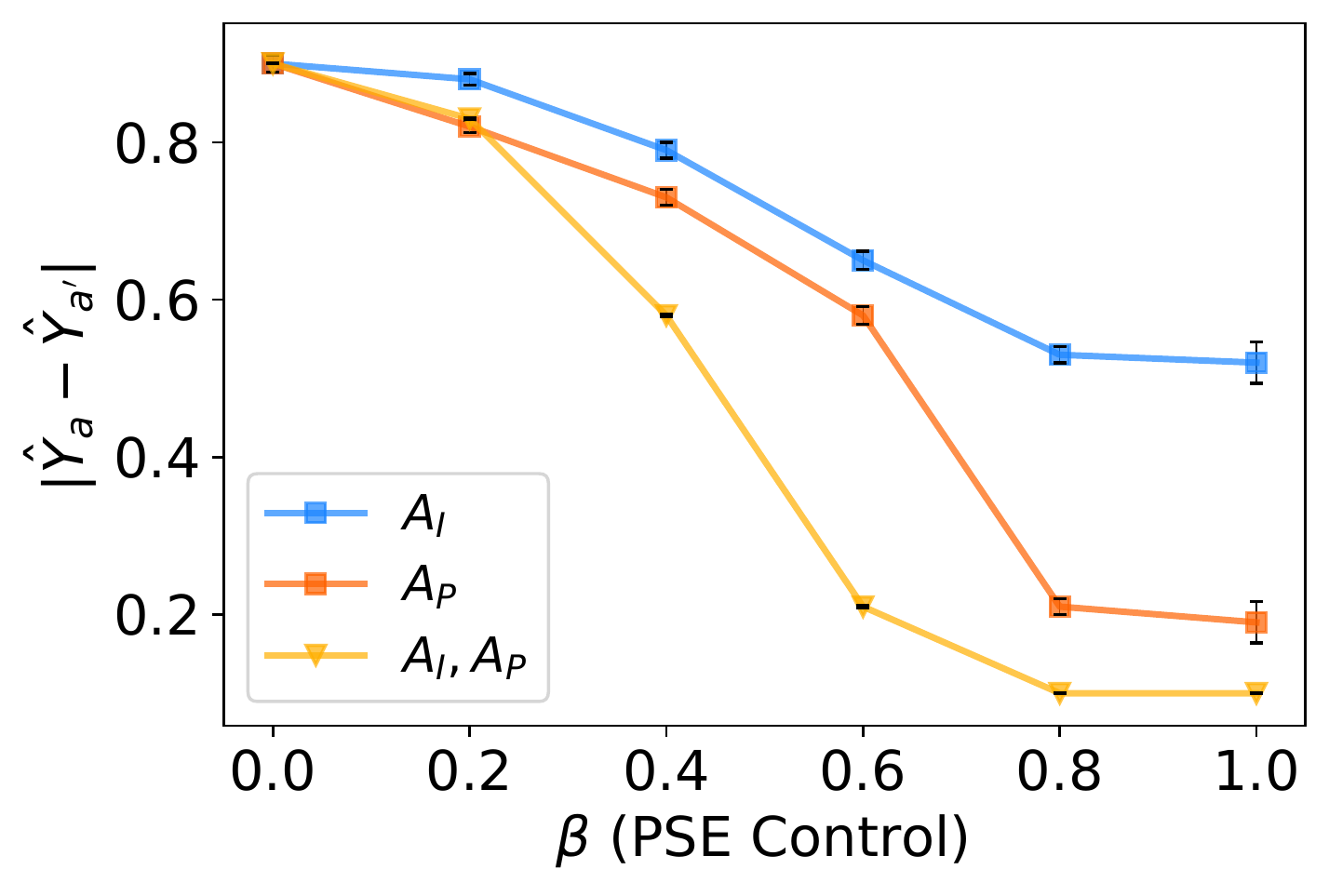}
    \caption{}
    \label{fig:uci_quant_unfair}
    \end{subfigure}
    \hfill
    \label{fig:real_world}
    \caption{a) Causal graph for the UCI Adult dataset \citep{chiappa2019path}, $A$ and $M$ represent the individual-level protected attributes, sex and marital status, respectively, $C$ is age and nationality, $L$ is the level of education, $R$ corresponds to working class, occupation, and hours per week, $Y$ is the income class, unfair paths are represented in green b) Density of $\hat{Y}$, c) path-specific unfairness, $\vert \hat{Y}_a - \hat{Y}_{a'}\vert$ controlling for the effects of just $A_I$ (blue), $A_P$ (orange) and both $A_I,A_P$ (yellow).}
\end{figure*}

We demonstrate real-world evaluation of the proposed approach on the Adult dataset from the UCI repository \citep{lichman2013uci}. The UCI Adult dataset is amenable for demonstration of our method, given that it was used in \citet{chiappa2019path} to assess path-specific counterfactual fairness, and because there are variables oriented in a manner that create a potential multi-level scenario. We consider this well-studied setting \citep{chiappa2019path,nabi2018fair} wherein the goal is to predict whether an individual's income is above or below \$50,000. The data consist of age, working class, education level, marital status, occupation, relationship, race, gender, capital gain and loss, working hours, and nationality variables for 48842 individuals. The causal graph is represented in Figure \ref{fig:uci}. Consistent with \citet{chiappa2019path} and \citet{nabi2018fair}, we do not include race or capital gain and loss variables. Here, $A$ represents the individual-level protected attribute sex, $C$ is nationality, $M$ is marital status, $L$ is the level of education, $R$ corresponds to working class, occupation, and hours per week, $Y$ is the income class.
We make an observation that nationality, $C$, could be considered as a non-aggregate macro-level sensitive attribute. This is because it may affect social influences and/or resources available at the macro-level. For example, nationality affects both education levels and martial status of individuals. Moreover, we also note that marital status, $M$, can also be considered as an individual-level sensitive attribute, as we may not want to discriminate individuals based on their marital status \citep{joslin2015marital,reimers1985cultural}. Thus, $C$, a macro-level sensitive attribute affects $M$, an individual-level sensitive attribute and the relationship between $C$ and other variables in the graph mimics that of a multi-level scenario. Moreover, it may be reasonable to consider a setting in which it is  desirable to predict income, while also being fair to nationality, in order to understand the effect that nationality may have. We highlight that macro-properties such as nationality are included in datasets like UCI adult, however, correcting for any unfairness due to macro-level sensitive attributes has not been considered in previous fairness studies. We also emphasize that there may be other datasets that better express the need for macro and individual-level sensitive attributes, however we consider this dataset/setting in order to be consistent with previous work in causal fairness which has used the same dataset, though have only analyzed unfairness due to an individual-level sensitive attribute \citep{chiappa2019path}. Paths in Figure \ref{fig:uci} are thus defined based on mapping the figure from \citet{chiappa2019path} to the multi-level framework in Figure \ref{fig:multi_sensitive} with $C$ as the macro-level sensitive attribute and $A$, $M$ as the individual-level sensitive attributes.


To first evaluate the path-specific effect of a single individual-level sensitive attribute as done in prior work \citep{chiappa2019path}, we evaluate the effect of only $A$ (the individual level sensitive attribute considered in \citet{chiappa2019path}) along $A\rightarrow Y$ and $A\rightarrow M \rightarrow \cdots \rightarrow Y$ to be 3.716. We obtain a fair prediction as follows:
\begin{align*}
    \hat{y}_{\text{fair}}= \text{Bernoulli}\left(\sigma\left(\theta^y + \theta^y_c C + \theta^y_m\left(M - \theta^m_a\right) + \theta^y_l\left(L - \theta^l_m \theta^m_a\right)\right)\right)
\end{align*}
by removing the path specific effect of $A$ along a priori known discriminatory paths. The fair accuracy is 78.87\%, consistent with \citet{chiappa2019path}. \\



We next evaluate the path-specific effect of both the macro ($C$) and individual-level sensitive attributes ($A,M$)\footnote{We use pre-processed data from \citep{nabi2019learning}.}. In the counterfactual world, the individual receives advantaged macro and individual level treatment. Since, $C$ is mostly represented by the value \emph{39}, and $A$ is represented by 2 levels 0 and 1, worst to best, we consider the baseline values to be $c=39$ and $a=1$.
We compute the path-specific effect of $C$ on $Y$ via $C \rightarrow M \rightarrow \cdots \rightarrow Y$, $C\rightarrow L \rightarrow \cdots \rightarrow Y$, $A \rightarrow M \rightarrow \cdots \rightarrow Y$, and $A\rightarrow Y$ in the baseline scenario to be 10.65. As we do not completely remove the effect of $C$; only along certain paths as described above, we still retain important information about nationality, $C$, an informative feature. After correcting for the unfair effect of $C$, $A$, and $M$, the accuracy is 77.92\% ($\beta=1$), in comparison with an accuracy of 78.87\% from removing the unfair effect of $A$ alone. Thus, there is not significant drop in the model performance after correcting for the unfair effects of both $A,M$ and $C$.

Next, we study the residual unfairness in a series of  counterfactual settings which account for unfairness at individual, macro-level sensitive attributes and their combination. Results are presented in Figure \ref{fig:uci_densityplots}. If density of two plots coincide, then they are counterfactually fair, i.e. there is no difference in prediction based on unfair pathways. The \emph{purple} plot represents the fair model, $\hat{Y}_{\text{fair}}$, trained on the original data that corrects for the multi-level path-specific effects of $\{A,M,C\}$. 
In the process of learning the fair predictions, we learn the parameters for the model represented in Figure \ref{fig:uci} as illustrated in Algorithm \ref{alg:multi-level_causal_fairness} and then generate counterfactual samples from the observed distribution. We next assess fairness of models that account for varying fair/unfair pathways and variable types. First, we generate individual-level counterfactual data by only altering the individual-level sensitive attributes $A$ and $M$ along the unfair pathways, and train a model on the individual-level counterfactual data, we call this $\hat{Y}\left(a_i'\right)$. The \emph{red} curve represents the density plot of $\hat{Y}\left(a_i'\right)$ after controlling for the individual-level path-specific effect at $\beta = 1$. Similarly, we generate macro-level counterfactual data by only altering the macro-level sensitive attribute, $C$, along the unfair paths, $C\rightarrow M \rightarrow \cdots, C\rightarrow L \rightarrow \cdots $, and train a model on this macro-level counterfactual data, which is represented by $\hat{Y}(a_p')$ (\emph{green} curve). At last, we generate individual and macro-level counterfactual data, by altering $A,M$ and $C$ along all unfair paths, represented by green in Figure \ref{fig:uci}. In this case, the predictor, $\hat{Y}(a_i',a_p')$, is represented by the \emph{black} curve,
As can be seen from the general overlap of the \emph{purple} and \emph{black} curves, controlling for path-specific effects of both individual and macro-level sensitive attributes generates counterfactually fair models, while only controlling for the individual or macro variables doesn't provide the same density estimate as a fair model. Similarly, the residual unfairness is presented in Figure \ref{fig:uci_quant_unfair}, and controlling for the multi-level path-specific effects (yellow curve) results in a counterfactual fair model as the difference, $\vert \hat{Y}_{a} - \hat{Y}_{a'}\vert$ approaches zero. Thus, correcting the multi-level path specific effect does not marginally drop the model performance, while ensuring that predictions are fair with respect to individual and macro-level attributes.

%% file: Figures/adult_uci.tex
\begin{tikzpicture}
    \node[state,red] (c) at (0,0) {C};
    \node[state] (y) [right =of c] {Y};
    \node[state] (r) [above =of y] {R};
    \node[state] (l) [above =of r] {L};
    \node[state,red] (m) [above =of l] {M};
    \node[state,red] (a) [above =of m] {A};
    \path (a) edge (m);
    \path [green] (m) edge (l);
    \path (l) edge (r);
    \path [green,dashed](l) edge (r);
    \path (r) edge (y);
    \path [green,dashed] (r) edge (y);
    \path (c) [green] edge [bend left=20] (m);
    \path (c) [green] edge (l);
    \path (c) edge (r);
    \path (c) edge (y);
    \path [green] (a) edge (m);
    \path (m) edge [bend left=30] (r);
    \path [green,dashed] (m) edge [bend left=30] (r);
    \path [green] (m) edge [bend left = 40] (y);
    \path (l)  edge [bend left = 30] (y);
    \path (l) [green,dashed]  edge [bend left = 30] (y);

    \path (a) edge [bend left = 30] (l);
    \path (a) edge [bend left = 30] (r);
    \path [green] (a) edge [bend left = 50] (y);


\end{tikzpicture}

%% file: discussion.tex
\section{Conclusion}
Our work extends algorithmic fairness to account for the multi-level and socially-constructed nature of forces that shape unfairness. In doing so, we first articulate a new definition of fairness, \emph{multi-level fairness}. Multi-level fairness articulates a decision being fair towards the individual if it coincides with the one in the counterfactual world where, contrary to what is observed, the individual identifies with the advantaged sensitive attribute at the individual-level and also receives advantaged treatment at the macro-level described by the macro-level sensitive attribute. A framework like this can be used to assess unfairness at each level, and identify the places for intervention that would reduce unfairness best (e.g. via macro-level policies versus individual attributes). As we show here, leveraging datasets that represent more than individual attributes can improve fairness, as often individual level attributes have variance with respect to macro-properties, or are proxies for macro-properties which may be the critical sources of unfairness \citep{mhasawade2020machine}. This is becoming possible given the large number of open data sets representing macro-attributes that are available, relevant to health, economics, and many other settings. Through our experiments, we illustrate the importance of accounting for macro-level sensitive attributes by exhibiting residual unfairness if they are not accounted for. Importantly, we show this residual unfairness persists even in cases when the same information and variables are considered, but without accounting for the multi-level nature of their interaction.

Finally, we demonstrate a method for achieving fair outcomes by removing unfair path-specific effects with respect to both individual and macro-level sensitive attributes. While a clear understanding of the macro-level sensitive attributes is essential for multi-level fairness, approaches to learning the latent representation of the macro-level sensitive attributes in their absence from individual-level factors could be important future work. As an initial framework, we consider path-specific effects for linear models here; we also endeavor to extend this to fewer assumptions on the functional form in the future.

%% file: appendix.tex
\section{Interaction between macro and individual-level variables}
\noindent \textbf{Causal effects in multi-level systems}
\label{cf_multi_level}
\begin{figure*}[!htbp]
\centering
\begin{subfigure}{.33\textwidth}
 \centering
    \scalebox{1.0}{\input{Figures/all_interactions}}
    \caption{ $I = f_I\left(\bf{P}\right),  
    Y = f_Y\left(\bf{I},\bf{P}\right).$}
    \label{fig:all_interactions}
\end{subfigure}%
\begin{subfigure}{.33\textwidth}
    \centering
    \scalebox{1.0}{\input{Figures/pop_only_indi}}
    \caption{  $I = f_I\left(\bf{P}\right),
    Y = f_Y\left(\bf{I}\right)$.}
    \label{fig:pop_indi}
\end{subfigure}%
\begin{subfigure}{.33\textwidth}
    \centering
    \scalebox{1.0}{\input{Figures/pop_only_clini}}
    \caption{$Y = f_Y\left(\bf{I},\bf{P}\right)$.}
    \label{fig:pop_clini}
\end{subfigure}
    \caption{Interactions between population variables $P$, individual variables $I$, and outcome $Y$, structural equations corresponding to the specific causal graphs are outlined for each of the structure in a), b), and c).}
    \label{fig:cases}
\end{figure*}

\noindent \textbf{Case 1.}  Both macro and individual level factors affect health outcomes through different causal paths. Figure \ref{fig:all_interactions} presents such a case where both individual-level factors and outcomes are affected by macro-level factors.
Recalling our definition of causal models, here $U = P, \textbf{V} = \{I,Y\}$. 
$P$ propagates its true value $p$ along the paths $P\rightarrow Y$ and $P\rightarrow I \rightarrow Y$. Therefore, we can estimate the causal effect of $P$ on $Y$ as follows:
\begin{align}\label{eq:both}
    \pr\left(Y\vert P = p \right) = \int_I\pr\left(Y\vert I,p\right)\pr\left(I\vert p\right).
\end{align}

\noindent \textbf{Case 2.} Macro-level factors affect outcomes indirectly by only affecting individual-level factors, and having no direct effect on the outcome. This scenario is represented in Figure \ref{fig:pop_indi}. 
In this setting the effect of $P$ on $Y$ is propagated only via $I$, i.e., via $P \rightarrow I \rightarrow Y$.
Thus the causal effect of $P$ on $Y$ is only an indirect effect, calculated as: 
\begin{align}\label{eq:mediate}
    \pr\left(Y\vert P = p\right) = \int_I \pr\left(Y\vert I\right) \pr\left(I\vert p\right).
\end{align}

\noindent \textbf{Case 3}. Macro and individual-level variables do not interact, but independently affect the outcome, as represented in Figure \ref{fig:pop_clini}. In this case $\pr \left(I\vert P=p\right) = \pr\left(I\right)$ since $I \independent P$. The causal effect of $P$ on $Y$ simplies to:
\begin{align}\label{eq:indep}
    \pr\left(Y\vert P = p\right) = \int_I \pr\left( Y \vert I, p\right)\pr\left(I\right).
\end{align}
and the structural equation is:
\begin{align*}
    Y &= f_Y\left(\bf{I},\bf{P}\right).
\end{align*}


\section{Evaluating path-specific effects independently for $A_P$ and $A_I$, respectively}
\noindent \underline{\textbf{ ($A_P$). }}We can use the same technique for evaluating the path-specific effect of $A_P=a_p$ along $\cdots \rightarrow I \rightarrow Y$ without intervening upon the individual level sensitive attribute $A_I$ as we did for identifying the effect of both $A_P$ and $A_I$. The counterfactual distribution in this case is as follows:
\begin{align}
        \int_{P,I,A_I} \pr\left(Y\vert A_I,I,P\right) \pr\left(I\vert A_I,a_p\right)\pr\left(P\vert a_p'\right)\pr\left(A_I\vert P,a_p'\right).
\end{align}
Here, we integrate with respect to $A_I$ since we are not interested in evaluating the effect of $A_I$ along the specified paths. Following the procedure from Section 5,  we obtain the path-specific effect along $\cdots \rightarrow I \rightarrow Y$ as:
\begin{align}
    \text{PSE}^{\cdots \rightarrow I \rightarrow Y}_{A_p \leftarrow a_p} = \theta^y_i \theta^i_{a_p}.
\end{align}

\noindent \underline{\textbf{($A_I$). }}We also assess the effect of only intervening on the individual-level attribute, $A_I$,  via paths $\cdots \rightarrow I \rightarrow Y$. The counterfactual distribution in this case looks as follows:
\begin{align}
    \int_{P,I,A_P} \pr\left(Y\vert a_i,I,P\right) \pr\left(I \vert a_i,A_P\right) \pr\left(P \vert A_P\right) \pr\left(A_P\right)
\end{align}
The mean distribution of which is:
\begin{align}
     \theta^y_{\phantom{i}} + \theta^y_i \theta^i_{\phantom{i}} + \theta^y_p \theta^p_{\phantom{i}} + \theta^y_{a_i} a_i' + \theta^y_i \theta^i_{a_i}a_i + \theta^y_i\theta^i_{a_p}a_p' + \theta^y_p \theta^p_{a_p} a_p'.
\end{align}
The path-specific effect along $\cdots \rightarrow I \rightarrow Y$ can thus be calculated as:
\begin{align}
\begin{split}
     &\theta^y_{a_i} \left(a_i'-a_i'\right) + \theta^y_i \theta^i_{a_i}\left(a_i-a_i'\right)\\ + &\theta^y_i\theta^i_{a_p}\left(a_p'-a_p'\right) + \theta^y_p \theta^p_{a_p} \left(a_p'-a_p'\right) \\
     =& \theta^y_i \theta^i_{a_i}\left(a_i-a_i'\right).
\end{split}
\end{align}
After substituting $a_i=1$ and $a_i' =0$ this results in:\\ $PSE_{A_i\leftarrow a_i}^{\cdots \rightarrow I \rightarrow Y} = \theta^y_i\theta^i_{a_i}.$ 

Therefore, the path-specific effect representing the unfairness along a path changes, based on which sensitive attributes were intervened upon, affecting the values of $a_i$ and $a_p$ in the counterfactual distribution.
Thus, the path-specific effects demonstrating unfairness
along the same sub path ($\dots \rightarrow I \rightarrow Y$) differ based on which variables (macro/individual-level sensitive attributes) are intervened upon.

\section{Path-specific effects for synthetic setting}
\begin{table*}[!htbp]
    \centering
    \begin{tabular}{l l l}
    \toprule
    Sensitive Attribute & Causal Path(s) & PSE\\
    \midrule
    \multirow{6}{*} {$A_P$} & $A_P \rightarrow P \rightarrow Y$ & $\theta^y_p\cdot \theta^p_{a_p}$\\
         &  $A_P \rightarrow I \rightarrow Y $ & $\theta^y_i\cdot\theta^i_{a_p}$\\
         & $A_P \rightarrow A_I \rightarrow I \rightarrow Y $ & $\theta^y_i \cdot \theta^i_{a_i} \cdot \theta^{a_i}_{a_p}$\\
         & $A_P \rightarrow A_I \rightarrow Y $ & $\theta^y_{a_i}\cdot \theta^{a_i}_{a_p}$\\
         & $A_P \rightarrow A_I \rightarrow Y $ & $\theta^y_{a_i} \cdot \theta^{a_i}_{a_p}$\\
         & $A_P \rightarrow \cdots \rightarrow I \rightarrow Y $ & $\theta^y_{i} \left(\theta^i_{a_p} + \theta^i_{a_i}\cdot \theta^{a_i}_{a_p}\right)$ \\
         & $A_P \rightarrow \cdots \rightarrow Y $ & $\theta^y_{p}\cdot \theta^p_{a_p} + \theta^y_{i} \left(\theta^i_{a_p} + \theta^i_{a_i}\cdot \theta^{a_i}_{a_p}\right) + \theta^y_{a_i}\cdot \theta^{a_i}_{a_p}$\\
         \midrule
         \multirow{3}{*}{$A_I$} & $A_I \rightarrow Y$ & $\theta^y_{a_i}$\\
                                & $A_I \rightarrow I \rightarrow Y$ & $\theta^y_i\cdot \theta^i_{a_i}$ \\
                                &  $ A_I \rightarrow \cdots \rightarrow Y$ & $\theta^y_{a_i} + \theta^y_{i} \cdot \theta^i_{a_i}$\\
        \midrule
        \multirow{7}{*}{$A_P,A_I$} & $A_P \rightarrow P \rightarrow Y, A_I \rightarrow Y$ & $\theta^y_p\cdot \theta^p_{a_p} + \theta^y_{a_i}$\\
        & $A_P \rightarrow P \rightarrow Y, A_I \rightarrow I \rightarrow Y$ & $\theta^y_p\cdot \theta^p_{a_p} + \theta^y_i \cdot \theta^i_{a_i}$\\
        & $A_P \rightarrow P \rightarrow Y, A_I \rightarrow I \rightarrow Y, A_I \rightarrow Y$ &  $\theta^y_p\cdot \theta^p_{a_p} + \theta^y_i \cdot \theta^i_{a_i} + \theta^y_{a_i}$\\
        & $A_P \rightarrow I \rightarrow Y, A_I \rightarrow I \rightarrow Y $ & $\theta^y_{i} \left(\theta^i_{a_i} + \theta^i_{a_p} \right)$\\
        & $A_P \rightarrow I \rightarrow Y, A_I \rightarrow Y$ & $\theta^y_{a_i} + \theta^y_i \cdot \theta^i_{a_p}$\\
        & $A_P \rightarrow I \rightarrow Y, A_I \rightarrow \cdots \rightarrow Y$ & $\theta^y_{a_i} +\theta^y_{i} \left(\theta^i_{a_i} + \theta^i_{a_p} \right) $\\
        & $A_P \rightarrow \cdots \rightarrow Y, A_I \rightarrow \cdots \rightarrow Y$ & $\theta^y_{a_i} +\theta^y_{i} \left(\theta^i_{a_i} + \theta^i_{a_p} \right) +\theta^y_i\cdot\theta^i_{a_p} $\\
        
        \bottomrule
    \end{tabular}
    \caption{Path-specific effects along different causal paths, each obtained by same procedure outlined in Section 5.}
    \label{tab:pse}
\end{table*}


\section{Experiments}
\subsection{Synthetic setting}
The \emph{blue} line represents the performance with respect to the control parameter ($\beta$) for PSE while only accounting for the path-specific effect of $A_I$, the \emph{orange} line represents performance while controlling for the path-specific effect of both $A_P$ and $A_I$, and the \emph{yellow} line represents the accuracy while controlling for the effect of $A_P$ alone, all along the path $\cdots \rightarrow I \rightarrow Y$. Since the PSE for $A_I$ is small (0.007) there is little effect on the model performance shown by the steady nature of the \emph{blue} line.
  \begin{figure}
    \includegraphics[scale=0.5]{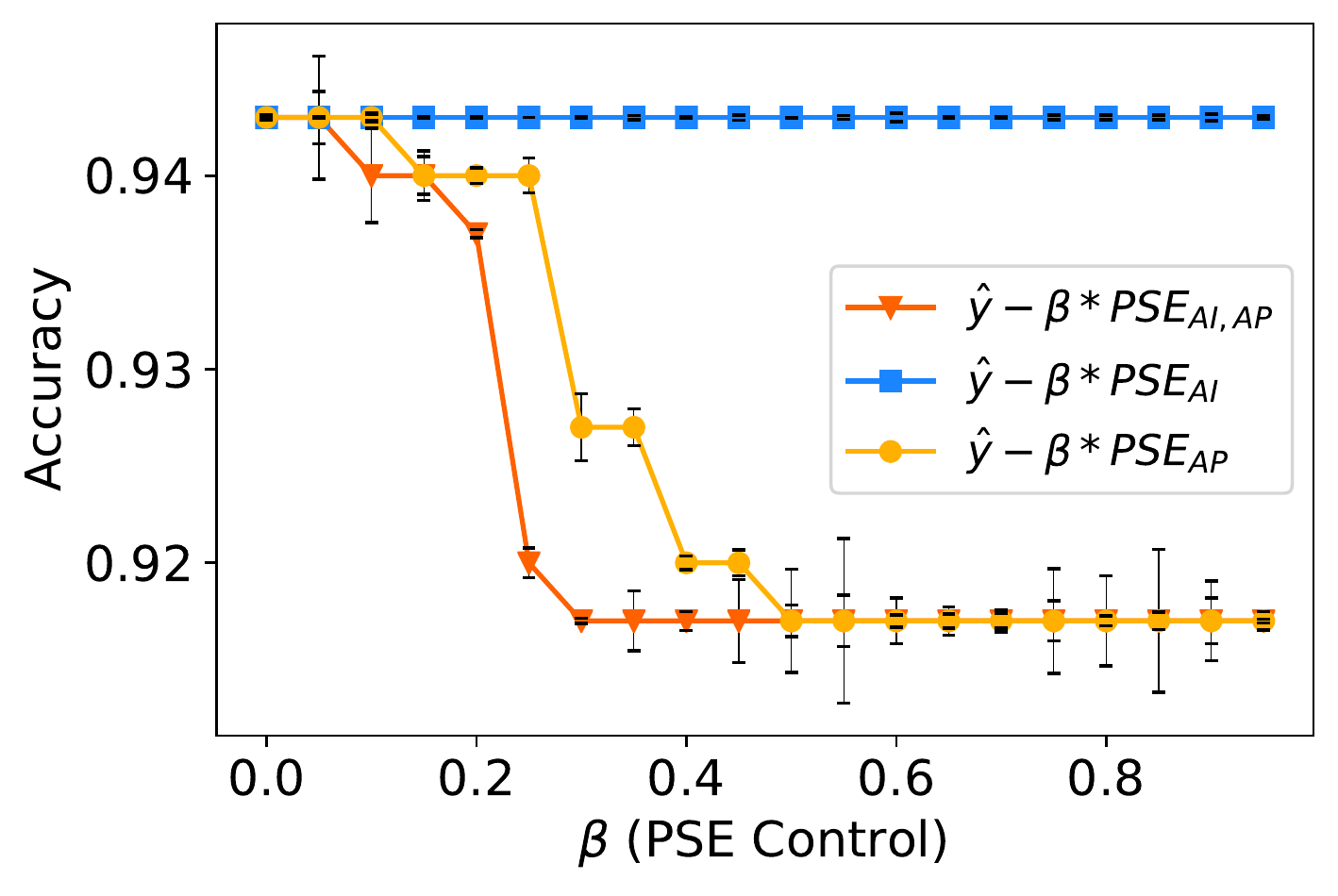}
    \caption{Model performance for different fair predictors, $\hat{y}_{\text{fair}}= \hat{y}-\beta*\text{PSE} $ accounting for PSE due to only $A_I$, $A_P$, and both $A_I,A_P$}
    \label{fig:result_acc}
    \end{figure}%
    
\section{Implementation details}
Classifiers (logistic regression and gradient boosting trees) are implemented using the package \texttt{scikit-learn} (v0.22.1) \citep{pedregosa2011scikit} in Python.
All experiments were ran on a single node of a compute cluster with a 3.0 GHz Intel processor and 8 GB memory.

%% file: Figures/all_interactions.tex
\begin{tikzpicture}
    \node[state] (p) at (0,0) {P};
    \node[state] (o) [right =of p] {Y};
    \node[state] (i) [below =of o] {I};
    \path (p) edge (i);
    \path (i) edge (o);
    \path (p) edge (o);
\end{tikzpicture}

%% file: Figures/pop_only_indi.tex
\begin{tikzpicture}
    \node[state] (p) at (0,0) {P};
    \node[state] (o) [right =of p] {Y};
    \node[state] (i) [below =of o] {I};
    
    \path (i) edge (o);
    \path (p) edge (i);
    
\end{tikzpicture}

%% file: Figures/pop_only_clini.tex
\begin{tikzpicture}
    \node[state] (p) at (0,0) {P};
    \node[state] (o) [right =of p] {Y};
    \node[state] (i) [below =of o] {I};

    \path (i) edge (o);
    \path (p) edge (o);
\end{tikzpicture}